\documentclass[runningheads]{llncs}

\usepackage{eccv}

\usepackage{eccvabbrv}

\usepackage[accsupp]{axessibility}  
\usepackage{hyperref}

\usepackage{orcidlink}

\usepackage[T1]{fontenc}
\usepackage[english]{babel}
\usepackage{graphicx}
\usepackage{amsmath}
\usepackage{amssymb}
\usepackage{gensymb}
\usepackage{nccmath} 
\usepackage{multirow}
\usepackage{bbding} 
\usepackage{caption}
\usepackage{booktabs}
\usepackage{stmaryrd} 
\usepackage{adjustbox} 
\usepackage{xcolor}
\usepackage{cite}
\usepackage{tikz}

\DeclareRobustCommand{\mkcircle}{\tikz[baseline=-0.5ex]\draw[black,fill=gray] (0,0) circle (.5ex);}
\DeclareRobustCommand{\mksquare}{\tikz[baseline=-0.5ex]\draw[black,fill=gray] (0,0) rectangle (1ex,1ex);}
\DeclareRobustCommand{\mktriangle}{\tikz[baseline=-0.5ex]\draw[black,fill=gray] (0,0) -- (1.2ex,0) -- (.6ex,1.2ex) -- cycle;}
\DeclareRobustCommand{\mkdiamond}{\tikz[baseline=-0.5ex]\draw[black,fill=gray] (.6ex,0) -- (1.2ex,.6ex) -- (.6ex,1.2ex) -- (0,.6ex) -- cycle;}

\newcommand{\preCR}[1]{{\color{red}#1}}
\renewcommand{\preCR}[1]{#1} 

\begin{document}

\title{Bringing BNNs to Fast Event Processing}
\titlerunning{Bringing BNNs to Fast Event Processing}

\author{Paul Longour\inst{1} \and
Julien Moreau\inst{1}\orcidlink{0000-0001-5008-9232} \and
Franck Davoine\inst{2}\orcidlink{0000-0002-8587-6997}}

\authorrunning{P. Longour et al.}

\institute{Université de technologie de Compiègne, CNRS, Heudiasyc, Compiègne, France \\
\email{\{paul.longour, julien.moreau\}@hds.utc.fr} \and
CNRS, INSA Lyon, LIRIS, UMR5205, Villeurbanne, France \\
\email{franck.davoine@cnrs.fr}}

\maketitle

\begin{abstract}
Binary Neural Networks (BNNs) enable efficient deep learning deployment on resource constrained devices with weights and activations compressed to one bit, substantially reducing model size and inference cost.
Event cameras offer complementary advantages, including low latency, high dynamic range, and low power consumption, by capturing asynchronous streams of events rather than dense image frames.
Despite their shared emphasis on efficiency, the combination of these technologies remains largely unexplored.
This work aims at adapting and evaluating modern deep BNN architectures on event data.
We also show that cross-modal pretraining from RGB data can improve the classification accuracy of BNNs on neuromorphic datasets.
We introduce the Polar-wise Binary Event Volume (PBEV), a binary representation that enables event-camera data to be processed directly by BNNs and represents a step toward fully binarized event-based vision systems.
Best evaluated BNN on N-Caltech101 classification benchmarks shows 90.58\% accuracy with 7.5$\times$ less operations than their full-precision counterparts.
\end{abstract}

\section{Introduction}
\label{sec:intro}

\begin{figure}[htb]
    \centering
    \includegraphics[width=0.7\textwidth]{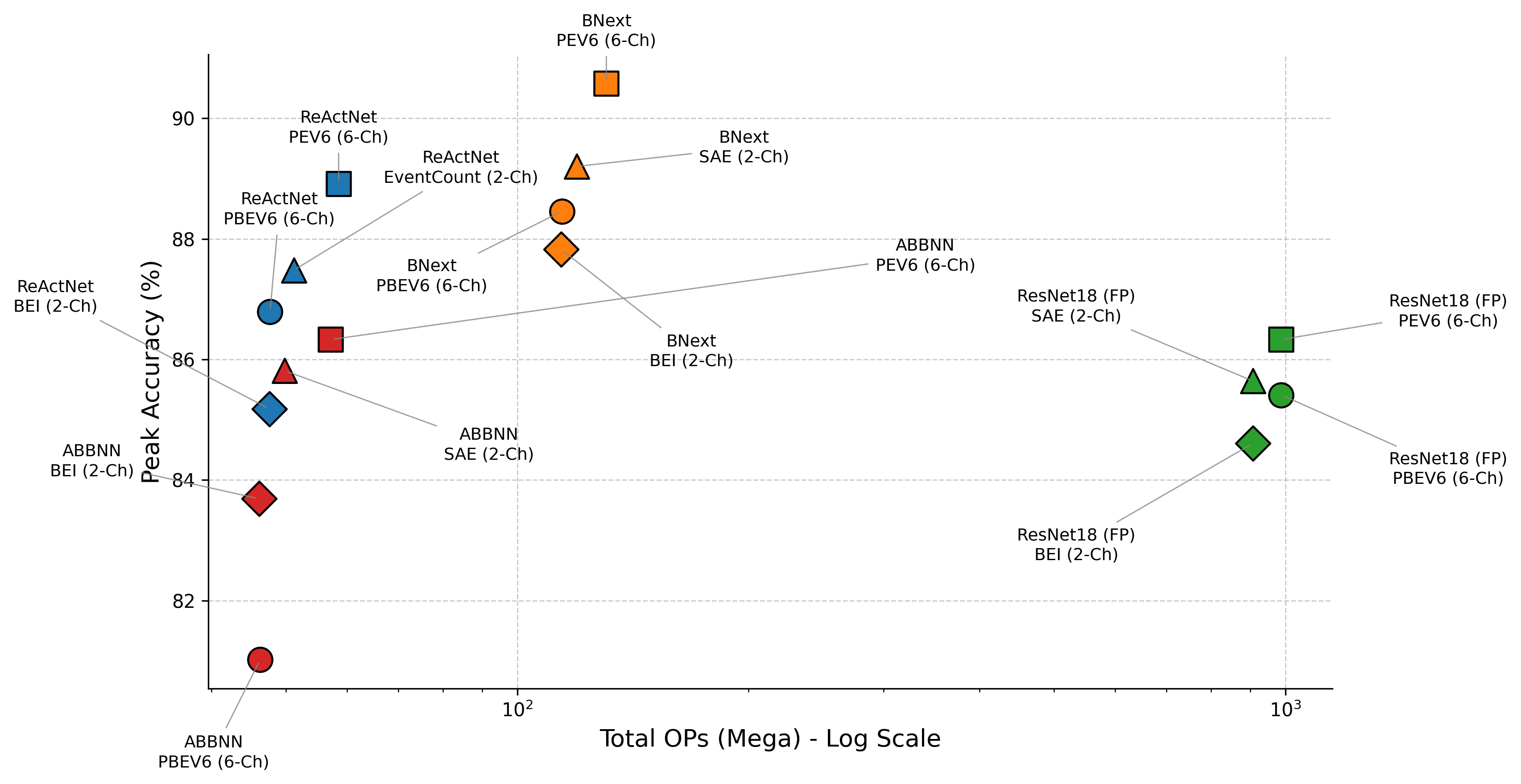}
    \caption{Classification accuracy on N-Caltech101~\cite{Orchard15} vs. Total Operations. Markers for input representations: \mktriangle~Best non-bin. 2-Ch (SAE / EventCount), \mksquare~PEV6 (non-bin. 6-Ch), \mkdiamond~BEI (bin. 2-Ch), and \mkcircle~PBEV6 (bin. 6-Ch). Colors for model architectures: \textcolor[HTML]{1f77b4}{\textbf{blue}} for ReActNet, \textcolor[HTML]{ff7f0e}{\textbf{orange}} for BNext, \textcolor[HTML]{2ca02c}{\textbf{green}} for ResNet-18, and \textcolor[HTML]{d62728}{\textbf{red}} for A\&B~BNN.}
    \label{fig:acc_vs_ops}
\end{figure}

Event cameras, also known as neuromorphic cameras, enable new vision capabilities by producing asynchronous, sparse streams of events rather than dense image frames captured at fixed rates. Offering an ultra-low latency
and a high dynamic range, event cameras show robustness in scenes with varying lighting conditions while requiring lower power consumption by avoiding redundant data~\cite{doi:10.1126/scirobotics.aaz9712}.
Currently, there are two main ways to implement neural networks for event data.
First are Spiking Neural Networks (SNN), which address the asynchronous and spike nature of the event flow but require specific neuromorphic hardware for deployment.
Second are classic artificial and Convolutional Neural Networks (CNN), which rely on pseudo-frame generation from event accumulation, reducing the inherent ultra-low latency of the camera. 
Binary Neural Networks (BNN) offer lighter models and lower power consumption, suiting embedded inference speeds and better preserving the events' low-latency benefit.
While progress has been made on BNNs, it remains unclear how they perform compared to classical CNNs and SNNs on neuromorphic data.
We evaluate accuracy and computational load of state-of-the-art binary and full precision architectures on event camera classification datasets across a range of accumulation windows and event representations (summarized in Figure~\ref{fig:acc_vs_ops}).

Our contributions can be listed as follows:
\begin{itemize}
    \item We introduce a novel Polar-wise Binary Event Volume (PBEV) representation for events. It is an efficient way to represent an accumulation of events into a binary tensor for direct processing with binary models.
    \item We benchmark state-of-the-art deep convolutional BNNs on event-based classification through an extensive evaluation against a full-precision CNN baseline across multiple event representations and accumulation schemes, and position the results against literature CNN, SNN and transformer results on two datasets, N-Caltech101~\cite{Orchard15} and N-ImageNet-mini~\cite{kim2022nimagenetrobustfinegrainedobject}.
    \item We analyse how representation, accumulation time, and cross-modal pretraining impact learning and generalization.
    \item We highlight how all tested BNN architectures can equal and surpass the full-precision CNN baseline ResNet-18 on event dataset, while requiring from $7\times$ up to $21\times$ less operations.
\end{itemize}


\section{Related Work}
\label{sec:biblio}

\subsection{Advances in Deep Binary Neural Networks}

The primary appeal of these extremely compressed networks lies in their deployment efficiency and hardware compatibility, with real implementation on FPGA.
A practical application of the state-of-the-art BNext model was deployed on an FPGA, achieving 72.6 FPS~\cite{ldfbnext} inference for its small variant and an energy consumption below 19W (on ZCU102 FPGA).

\vspace{-1em}
\subsubsection{Deep BNNs}

To scale BNNs to complex classification datasets, modern architectures have relied on the same core breakthroughs that advanced standard CNNs.
For instance, residual connections~\cite{liu2018birealnetenhancingperformance} can propagate real-valued activations around binary convolution blocks, thereby mitigating the loss of representational capacity caused by binarization. Batch normalization~\cite{sari2020doesbatchnormalizationhelp, chen2021bnnbn} can improve optimization stability and control activation distributions during BNN training.
By adapting these classical deep learning techniques, along with BNN-specific refinements like learnable threshold optimization~\cite{liu2020reactnetprecisebinaryneural, lee2023instabnnbinaryneuralnetwork, liu2021adamtrainingstrategieshelp} and knowledge distillation methods tuned to reduce BNNs overfitting~\cite{guo2022joinhighaccuracyclub}, binary networks can now converge effectively on large-scale tasks.

Binarization has also been extended to transformer backbones, where the central difficulty is finding an accurate 1-bit encoding for the queries, keys and values inside attention~\cite{he2023bivitextremelycompressedbinary, le2023binaryvitpushingbinaryvision}.
Binarization has even reached language processing, with BiBERT~\cite{qin2022bibertaccuratefullybinarized} yielding a fully binarized BERT, and complete Large Language Models built on binary operations~\cite{bitnet}.

\vspace{-1em}
\subsubsection{Training BNN}
\label{sssec:sotatrainBNN}

A major bottleneck remains training: most BNNs require substantially longer training to converge. BNNs usually rely on a 2 steps training protocol~\cite{liu2018birealnetenhancingperformance}, the first one is to binarize only activations, then both activations and weights.\cite{liu2021adamtrainingstrategieshelp} explain why Adam optimizer outperforms stochastic gradient descent on BNNs.
The authors also study how training strategies affect the retention of the initial weight signs, using a correlation-to-initialization metric.
\emph{Knowledge distillation}~\cite{hinton2015distillingknowledgeneuralnetwork} is a training strategy for a smaller student network to match a larger teacher's soft output distribution rather than only the hard ground-truth labels, used in many BNN training pipelines.
While 1-bit activations limit feature map expressivity, they also act as implicit regularization~\cite{guo2022joinhighaccuracyclub}.


\subsection{Deep Learning Models to process Events}

Table~\ref{tab:sota} reports state-of-the-art classification results from events on N-Caltech101 and N-ImageNet datasets across various architecture types, namely CNN, Transformer, SNN, and Graph Convolutional Networks (GCN).

\begin{table}[htb]
\caption{State-of-the-Art Classification Performance on N-Caltech101~\cite{Orchard15} and N-ImageNet~\cite{kim2022nimagenetrobustfinegrainedobject}. For reported accuracy, Train/Val sets used by authors are not consistent.}
\label{tab:sota}
\centering
\begin{adjustbox}{width=1\textwidth}
\begin{tabular}{@{}llcccccll@{}}
\toprule
\multirow{2}*{\textbf{Architecture type}} & \multirow{2}*{\textbf{Method / Model}} & \multicolumn{2}{c}{\textbf{Accuracy (\%)}} & \multirow{2}*{\textbf{Timesteps}} & \multicolumn{2}{c}{\textbf{Processing Resolution}} & \multirow{2}*{\textbf{Backbone}} & \multirow{2}*{\textbf{Special training procedure}} \\
 & & \textbf{N-Caltech101} & \textbf{N-ImageNet} & & \textbf{N-Caltech101} & \textbf{N-ImageNet} & & \\
\midrule
\multirow{4}{*}{\textbf{CNN}}
 & EventDrop~\cite{eventdrop} &  87.14 & -- & -- & $180\times240$ & -- & MobileNet-V2 & Data augmentation \\
 & EventDrop~\cite{eventdrop} & 85.15 & -- & -- & $180\times240$ & -- & ResNet-34 & Data augmentation \\
 & ResNet-34~\cite{kim2022nimagenetrobustfinegrainedobject} & 68.12 / 80.88 / 86.81 & 48.43 / -- / -- & -- & $240\times304$ & $224\times224$ & ResNet-34 & -- / ImageNet pretrain / N-ImageNet pre. \\
\midrule
\multirow{5}{*}{\textbf{Transformer}} 
 & MEM~\cite{10484185} & 85.60 & 57.89 & -- & $224\times224$ & $224\times224$ & ViT & Self supervised encoder \\
 & MEM-NImNet~\cite{10484185} & 90.10 & -- & -- & $224\times224$ & -- & ViT & Self supervised after N-ImageNet pretrain \\
 & EventBind~\cite{zhou2023clip} & 95.29 & 63.54 & -- & $224\times224$ & $224\times224$ & ViT-L/14 & RGB-text-event CLIP contrastive learning \\ 
 & STP~\cite{STP} & 94.74 & 68.87 & -- & $224\times224$ & $224\times224$ & ViT-S/16 & Event to RGB domain alignment \\ 
 & GEP~\cite{cao2026generativeeventpretrainingfoundation} & 93.05 & 65.11 & -- & $224\times224$ & $224\times224$ & ViT-S/16 & DINOv2 contrastive learning and datasets fusion \\
 & GEP~\cite{cao2026generativeeventpretrainingfoundation} & 96.47 & 75.20 & -- & $224\times224$ & $224\times224$ & ViT-B/16 & DINOv2 contrastive learning and datasets fusion \\
\midrule
\multirow{7}{*}{\textbf{Spiking CNN}} 
 & NDA~\cite{li2022neuromorphicdataaugmentationtraining} & 83.70 & -- & 10 & $128\times128$ & -- & VGG-SNN & Data augmentation \\
 & EventMix~\cite{SHEN2023119170} & 79.47 & -- & 10 & $48\times48$ & -- & ResNet-18 SNN & Data augmentation \\
 & Knowledge Transfer~\cite{efficientknowledgetransfert} & 93.45 & -- & 10 & $48\times48$ & -- & VGG-SNN & RGB to event sliding training \\
 & SMA-AZO-VGG~\cite{shan2024advancingspikingneuralnetworks} & 84.60 & -- & 14 & $180\times240$ & -- & VGG-SNN & -- \\
 & CKD~\cite{CKD} & 97.13 & -- & 10 & $48\times48$ & -- & VGG-SNN & RGB to event sliding training \\ 
 & TMC~\cite{pmlrv267yan25c} & 86.03 / 88.24 & -- & 10 / 16 & -- & -- & VGG-SNN & -- \\
 & TMKT~\cite{xie2025breakingmodalitywalltimestep} & \textbf{97.93} & -- & 10 & $48\times48$ & -- & VGG-SNN & Mixed RGB-event for domain alignment \\
 \midrule
\textbf{Spiking Transformer} & SpiLiFormer~\cite{zheng2025spiliformer} & 89.18 & -- & 16 & $128\times128$ & -- & Spiking ViT & -- \\
 \midrule
\multirow{2}{*}{\textbf{GCN}} & AEGNN~\cite{Schaefer22} & 64.30 & -- & -- & $256^3$ & -- & SplineConv & -- \\
 & EFGCN~\cite{JEZIOREK2026103850} & 62.80 & -- & 50 ms & $256^3$ & -- & PointNetConv & -- \\
\bottomrule
\end{tabular}
\end{adjustbox}
\end{table}

\vspace{-1em}
\subsubsection{Neuromorphic Learning}

On suitable event-driven hardware, SNNs can replace many dense multiply–accumulate (MAC) operations with conditional synaptic accumulations triggered by spikes.
Deep spiking CNNs like TMC~\cite{pmlrv267yan25c} adds temporal calibration to increase temporal gradient diversity.
SpiLiFormer~\cite{zheng2025spiliformer} introduces a lateral inhibition mechanism into spiking transformers to reduce attention distraction towards irrelevant tokens.

To benefit from the huge quantity of static RGB images, many SNNs rely on \emph{transfer learning} strategies.
Works like~\cite{efficientknowledgetransfert} apply domain sliding training: training the spiking backbone initially on RGB data then progressively shifting the input distribution toward events, so that RGB learned representations carry over into the event data domain.
Building on this mechanism, CKD~\cite{CKD} additionally distills logits from an RGB teacher.
TMKT~\cite{xie2025breakingmodalitywalltimestep} interpolates RGB and event inputs as a curriculum to bridge the modality gap for domain alignment.

Whether trained purely on events or from RGB, all of these networks remain spiking at inference and share the same bottleneck: the evaluated configurations commonly use 10–16 timesteps~\cite{li2022neuromorphicdataaugmentationtraining, SHEN2023119170, zheng2025spiliformer, pmlrv267yan25c}, which increases the number of sequential updates and may raise latency and computational cost.

\vspace{-1em}
\subsubsection{Full precision models}

To compensate for the scarcity of large-scale event datasets, the most performing methods adapt vision transformers with feature alignment from reference models trained on internet scale image data.
EventBind~\cite{zhou2023clip} binds event, image and text embeddings for \emph{contrastive learning} into a shared
CLIP~\cite{pmlr-v139-radford21a} space.
EventBind reliance on a large CLIP ViT backbones may limit its suitability for resource constrained deployment.
With even more compute resources, GEP~\cite{cao2026generativeeventpretrainingfoundation} uses a similar technique to align an event encoder to a RGB foundation model (DINOv2) before training on the Event-1.8M, a dataset with 1.8M event-image pairs drawn from 
various representative neuromorphic datasets.
These methods report among the highest published accuracies under their respective evaluation protocols (Table~\ref{tab:sota}), but largely by inheriting the parameter count and pretraining cost of large RGB foundation models.

Other methods apply \emph{transfer learning}, through simple weights pretraining, or domain alignment technique.
Pretraining is used in~\cite{kim2022nimagenetrobustfinegrainedobject, 10484185}, with~\cite{kim2022nimagenetrobustfinegrainedobject} applying also cross-modal pretraining from RGB.
Domain alignment is adopted by STP~\cite{STP}, who freezes a RGB pretrained backbone and adds modules specialized for events to make the features compatible.

In addition, MEM~\cite{10484185} trains a variational autoencoder and employs a masked patches technique to self-supervise a ViT backbone for events from unlabeled data, before fine-tuning for classification.




GCNs have also been proposed to process events~\cite{Schaefer22, JEZIOREK2026103850}.
They process the sparse flow in a spatio-temporal graph where each event is a vertex, avoiding a lossy compression to a pseudo-frame.
Despite this advantage, processing a graph is heavier than classic networks layers, and GCNs are not as deep as CNNs.

\subsection{Bridging BNNs and Event Vision}

Despite the potential compatibility between binarized event representations and 1-bit neural architectures, the application of BNNs to neuromorphic vision is largely unexplored.
Existing literature is limited to low-power applications such as 3-class parking lot monitoring~\cite{alwayson} and denoising for pedestrian detection~\cite{9191148}.
Prior binary event representations such as the Binary Event Image (BEI)~\cite{8260859} and the Binary Event History Image (BEHI)~\cite{Wang22} binarize the input representation, but the subsequent neural processing is not necessarily fully binarized.
We hypothesize that because event data natively represents event/no event per pixel and polarity, it can be encoded directly into dense binary formats.
This lets a BNN ingest space-time event volumes natively,
enabling the convolutional core to rely predominantly on XNOR–popcount operations.

\section{Methodology}
\label{sec:method}



\subsection{Datasets}
\label{sec:dataset}

\preCR{Event datasets originate either from live recordings (unbiased but rare and task-specific), from frame-to-event simulation (limited by the source camera's dynamic range and frame interpolation noise), or from pan-tilt recordings of static images, which conveniently produce large annotated datasets with a real sensor at the cost of true object dynamics.}

Here we use pan-tilt recordings that reproduce large and influent computer vision datasets.
N-Caltech101~\cite{Orchard15} is the event-based translation of Caltech101~\cite{FEIFEI200759} while N-ImageNet~\cite{kim2022nimagenetrobustfinegrainedobject} is from ImageNet~\cite{ILSVRC15}, N-ImageNet-mini is a 100-class subset from it.
Datasets used in our benchmark are detailed in Table~\ref{tab:dataset}.



\begin{table}[htb]
\caption{Summary of used datasets. Occupancy ratio is defined as the average ratio of pixels triggering at least once during the full sequence over the number of pixels.}
\label{tab:dataset}
\centering
\begin{adjustbox}{width=1\textwidth}
\begin{tabular}{lccccccc}
\hline
\textbf{Dataset} &
\textbf{Sample length} &
\textbf{Resolution} &
\textbf{\# Classes} &
\textbf{\# Train} &
\textbf{\# Val} &
\textbf{Occupancy} &
\textbf{Mean \# events} \\
\hline
N-ImageNet-mini~\cite{kim2022nimagenetrobustfinegrainedobject} & $50\,\mu s$ & $640\times480$ & 100 & 129{,}496 & 5{,}101 & 0.22 & 103{,}911 \\
N-Caltech101~\cite{Orchard15}    & $300\,ms$   & $240\times180$ & 101 & 7{,}000 & 1{,}709 & 0.41 & 115{,}610  \\
\hline
\end{tabular}
\end{adjustbox}
\end{table}

\begin{figure}[htb]
    \centering
    \begin{minipage}[t]{0.18\linewidth}\centering
        \includegraphics[width=\linewidth]{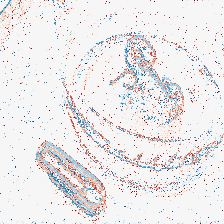}\\N-ImageNet-mini \preCR{(full 50\textmu{}s)}
    \end{minipage}\hfil
    \begin{minipage}[t]{0.18\linewidth}\centering
        \includegraphics[width=\linewidth]{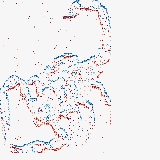}\\N-Caltech101 (5ms)
    \end{minipage}\hfil
    \begin{minipage}[t]{0.18\linewidth}\centering
        \includegraphics[width=\linewidth]{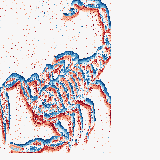}\\N-Caltech101 (50ms)
    \end{minipage}\hfil
    \begin{minipage}[t]{0.18\linewidth}\centering
        \includegraphics[width=\linewidth]{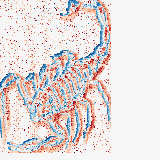}\\N-Caltech101 (\preCR{full} 300ms)
    \end{minipage}
    \caption{SAE representation (positive/negative channels merged into red/blue) of same-class ``scorpion'' samples from N-ImageNet-mini and N-Caltech101.}
    \label{fig:sae_cross_dataset}
\end{figure}

These datasets have different characteristics.
Although they contain approximately the same mean total number of events, N-ImageNet's sensor has roughly seven times more pixels and samples are approximately 6,000 times shorter than N-Caltech101 ones, resulting in a lower occupancy ratio.
N-Caltech101 is made to be processed with long accumulation times, it empirically provides too few events to be processed with durations below 5 ms.
When observing accumulated pseudo-frames, N-ImageNet presents sparse and thin edges due to its very short sample length while N-Caltech101 provides denser renderings~\ref{fig:sae_cross_dataset}.
The very short duration of N-ImageNet samples makes this dataset very challenging, but also a step to solve for low latency events processing with few aggregated events.

\subsection{Evaluated Architectures}
\label{ssec:architectures}

We select three state-of-the-art deep convolutional BNNs originally applied to images and one full-precision CNN baseline for our comparisons.
Following the knowledge distillation procedure to train BNNs (see section~\ref{sssec:sotatrainBNN}), we adopt EventBind~\cite{zhou2023clip}, a state-of-the-art full-precision transformer, as the teacher.
The networks are listed as follows and their complexity is reported
in section~\ref{ssec:complexity-models}.

\begin{itemize}
    \item \textbf{ReActNet~\cite{liu2021adamtrainingstrategieshelp}:} Built on the MobileNetV1~\cite{howard2017mobilenetsefficientconvolutionalneural} pattern (without depthwise separable convolutions), it matches ResNet-18 on ImageNet (70.5\%) at $22\times$ fewer operations.
    \item \textbf{A\&B~BNN~\cite{ma2024abbnnaddbitoperationonlyhardwarefriendly}:} Designed strictly for hardware supporting only bitwise operations, attaining on ImageNet 66.89\% accuracy. It builds upon BNN-BN~\cite{chen2021bnnbn} to fix bottlenecks related to heavy batch normalization.
    \item \textbf{BNext~Small~\cite{guo2022joinhighaccuracyclub}:} A more recent BNN reaching 76.1\% accuracy on ImageNet. This model has been deployed into a FPGA at 72.6 FPS~\cite{ldfbnext}.

    \item \textbf{ResNet-18~\cite{he2015deepresiduallearningimage}:} Evaluated as a standard, lightweight full-precision CNN baseline with 69.8 \% accuracy on ImageNet.
%
    \item \textbf{EventBind~\cite{zhou2023clip}:} We employ ViT-L/14 backbone, used as the teacher model for knowledge distillation.
    \preCR{As authors weights are not available,} we trained it ourself and achieve 94.08\% accuracy on the N-Caltech101 and 63.54\% on N-ImageNet-mini validation split.
\end{itemize}

\subsection{Polar-wise Binary Event Volume Representation}
\label{ssec:bev}

In recent BNN architectures, the first layer is kept in full precision.
As a result, the number of floating-point operations (FLOPs) increases with the number of input channels.
We adapt both the input representation and the network architectures by replacing this real-valued first layer with a binary layer, yielding binary-input variants of each network.
This reduces 
computational overhead while leaving the remainder of the architecture unchanged.
This design limits the increase in computational cost associated with higher-dimensional inputs. 

To enable this fully binary input processing, we introduce the Polar-wise Binary Event Volume (PBEV) encoding.
It consists of a stack of short time-slices formed into a 3D tensor with each polarity considered separately, similar to the polar-wise variant~\cite{Perot20} of Event Volume~\cite{Zhu19} \preCR{(PEV)}.
In PBEV, each slice is coded strictly as a binary channel representing the presence or absence of events at a given position. 
Unlike a 2D \preCR{BEI},
the PBEV \preCR{sub-samples the accumulation time window and} preserves most of the temporal density of the event flow.

Let $T$ be the full accumulation time, $C$ the total number of channels, and $e(x,y,t,p)$ the definition of an event at pixel coordinates $(x,y)$. Event's timestamp $t$ is expressed relative to the observation window, such that $t \in [0, T]$, while $p \in \{-1, +1\}$ denotes event's polarity. 
For $i \in \llbracket 0, \tfrac{C}{2}-1 \rrbracket$, the $i^{\text{th}}$ channel $I$ and $(i+\tfrac{C}{2})^{\text{th}}$ channel $J$ of the PBEV are defined as:
\vspace{-0.5em}
\begin{align*}
 \medmath{\forall (x,y) \in I, I(x,y)} &\medmath{= \left( \sum_{t = i \cdot \frac{2T}{C}}^{(i+1) \cdot \frac{2T}{C}} e(x,y,t,p = +1) \right) > 0,} \\
 \medmath{\forall (x,y) \in J, J(x,y)} &\medmath{= \left( \sum_{t = i \cdot \frac{2T}{C}}^{(i+1) \cdot \frac{2T}{C}} e(x,y,t,p = -1) \right) > 0}
\end{align*}


\subsection{Complexity Computation}
\label{ssec:complexity-models}

\begin{table}[htb]
\caption{Comparison of network architectures in terms of parameters and operations, at $160\times160$ input resolution. All quantities are reported in Mega (M).
Bin. input means binary input together with binarized convolutional layers.
}
\label{tab:op-mini}
\centering
\begin{adjustbox}{width=1\textwidth}
\begin{tabular}{lcccccc}
\toprule
\textbf{Architecture} & \preCR{\textbf{Layers}} & \textbf{Channels} & \textbf{Params (M)} & \textbf{BOPs (M)} & \textbf{FLOPs (M)} & \textbf{Total OPs (M)}\\
\midrule
EventBind (\preCR{ViT-L/14}) & \preCR{36} & 3 & 902.26 & - & $10{,}416{,}360$ & $10{,}416{,}360$ \\
ResNet-18 & \preCR{18} & 2 / 6 & 11.23 / 11.24 & - & 907.32 / 987.60 & 907.32 / 987.60 \\
\midrule
BNext~Small & \preCR{28} & 2 / 6 & 67.21 / 67.21 & $5{,}530$ & 33.17 / 44.23 & 119.57 / 130.63 \\
ReActNet & \preCR{28} & 2 / 6 & 28.42 / 28.42 & $2{,}460$ & 12.77 / 20.14 & 51.17 / 58.54 \\
A\&B~BNN & \preCR{28} & 2 / 6 & 28.41 / 28.41 & $2{,}460$ & 11.38 / 18.75 & 49.78 / 57.15 \\
\midrule
BNext~Small (bin. input) & \preCR{28} & 2 / 6 & 67.21 / 67.21 & $5{,}540$ / $5{,}550$ & 27.64 & 114.13 / 114.30 \\
ReActNet (bin. input) & \preCR{28} & 2 / 6 & 28.42 / 28.42 & $2{,}460$ / $2{,}470$ & 9.08 & 47.54 / 47.66 \\
A\&B~BNN (bin. input) & \preCR{28} & 2 / 6 & 28.41 / 28.41 & $2{,}460$ / $2{,}470$ & 7.69 & 46.15 / 46.27 \\
\bottomrule
\end{tabular}
\end{adjustbox}
\end{table}

Network complexity is linked to the quantity of parameters and of operations.
Operations OPs~\cite{martinez2020trainingbinaryneuralnetworks} combine binary and floating-point operations (FLOPs) following this formula: $OPs = BOPs/64 + FLOPs$.
\preCR{As shown in Table~\ref{tab:op-mini}, binarizing the input layer makes a 6-channel binary-input variant cheaper than its own 2-channel full-precision-input counterpart (\textit{e.g.} 48 vs. 51 M~OPs for ReActNet), leaving the final linear layer as the only remaining source of real multiplications.}


A\&B~BNN is built on top of ReActNet, having approximately the same number of parameters but replaces the multiplication operations with
more efficient bit operations~\cite{ma2024abbnnaddbitoperationonlyhardwarefriendly}: it drops batch normalization in favor of scaled weight standardization, absorbs the scaling activation multiplications into the sign function at inference, and constrains its PReLU slopes to integer powers of 2.
For comparability, Table~\ref{tab:op-mini} counts these shift and addition operations as regular FLOPs, so the reported A\&B~BNN totals are a loose upper bound.

BNNs efficiency stands in contrast to state-of-the-art full-precision models like EventBind ViT-L/14, which demands over 902 M~parameters and a staggering 10.4\,TFLOPs.
Even compared to advanced binary networks like BNext~Small, which requires 67.21 M~parameters and 120 M~OPs.


We do not compare BNNs to SNNs complexity in this work, as most SNN reduce drastically the input resolution (see Table~\ref{tab:sota}), lowering the number of parameters and operations.
Also, the theoretical Synaptic Operations (SOP) proposed in~\cite{SOP} depends on the neurons firing rate, which is time-dependent.

\section{Experiments}
\label{sec:experiments}

To assess the effectiveness of BNNs on event data, we benchmarked on the N-Caltech101~\cite{Orchard15} and N-Imagenet-mini~\cite{kim2022nimagenetrobustfinegrainedobject} classification datasets against the several architectures of Section~\ref{ssec:architectures} representing different efficiency and capability tradeoffs. Beyond raw accuracy, our experiments aim to evaluate the adaptability of these models across varying event representations and accumulation times.

\subsection{Evaluation metrics}

We use the top-1 accuracy results to compare performance and compute the correlation-to-initialization (C2I) ratio for measuring the dependency between trained parameters and initialization as in~\cite{liu2021adamtrainingstrategieshelp}. 
The C2I ratio is defined as:
\vspace{-1.5em}
\begin{center}
\[
    \text{C2I} = 1 - \frac{\sum_{l=1}^{L} \sum_{w \in W_l} \text{I}_{\text{C2I}}}{N_{\text{total}}}, \text{with }
    \text{I}_{\text{C2I}} = \frac{|\text{Sign}(w_{\text{final}}) - \text{Sign}(w_{\text{init}})|_{abs}}{2}
\]
\end{center}
\noindent where $N_{\text{total}}$ is the total number of binary weights, and $w_{\text{final}}$, $w_{\text{init}}$ the final and initial weights, respectively.


\subsection{Event Representations}
\label{ssec:event_repr}

We compare four families of event encoders: density-based 2D histograms~\cite{Maqueda_2018}, time-based Surfaces of Active Events~\cite{Alzugaray18, Park16}, 3D Event Volumes (or Voxels)~\cite{Perot20}, and binary representations as Binary Events presence Image~\cite{8260859} and proposed Polar-wise Binary Event Volume. The specific representations and configurations compared in our experiments are detailed in Table~\ref{tab:ev-repr}.
\preCR{Since binary convolutions are far cheaper than real-valued ones
(Section~\ref{ssec:complexity-models}), PBEV's channel count can be scaled at
minimal compute cost, enabling wide inputs with fast inference, whereas
real-valued inputs scale poorly. We use 6 channels as a representative
multi-bin setting, matched by the non-binary PEV6 for a direct comparison.}

\begin{table}[htb]
\caption{Comparison of Event Data Representations and Settings. Binary representations are boldened. P, T, and D denote the encoding of Polarity, Time, and Density, respectively. Parentheses (\checkmark) indicate partial or discretized encoding. A PBEV formulated with 2 channels is equivalent to the Binary Event Image.}
\label{tab:ev-repr}
\centering
\begin{adjustbox}{width=1\textwidth}
\begin{tabular}{l c l c c c}
\toprule
\textbf{Name} & \textbf{Channels} & \textbf{Description} & \textbf{P} & \textbf{T} & \textbf{D} \\
\midrule
2D Histogram (EventCount)~\cite{Maqueda_2018} & 2 & 2D histogram of event counts, + and - separated & \checkmark & & \checkmark \\
Surface of Active Events (SAE)~\cite{Alzugaray18} & 2 & Time surface retaining most recent timestamps, + and - separated & \checkmark & (\checkmark) & \\
Polar-wise Event Volume (PEV)~\cite{Perot20} & 6 & Time decayed events in 3D voxels, + and - separated & \checkmark & \checkmark & (\checkmark) \\
\textbf{Binary Event Image (BEI)}~\cite{8260859} & 2 & Boolean presence of events, + and - separated & \checkmark & & \\
\textbf{Polar-wise Binary Event Volume (PBEV)} & 6 & Event presence in 3D voxels, + and - separated (Section~\ref{ssec:bev}) & \checkmark & (\checkmark) &  \\
\bottomrule
\end{tabular}
\end{adjustbox}
\end{table}

\subsection{Implementation details}


\subsubsection{Data temporal sampling}
\label{ssec:dataset_repr}

N-Caltech101 samples are $300$\,ms long. To probe the accuracy--latency trade-off we accumulate events over windows of $5$, $10$, $20$, $50$, $100$, $200$ and the full $300$\,ms. For a window shorter than the sample, events are taken from a sliding temporal window: at training time its start is sampled uniformly around the sample centre, providing an extra layer of temporal augmentation, whereas at test time the window is centered on the middle timestamp. The dataset is resized to $160\times160$ following~\cite{zubic2023chaoscomesorderordering}.
We use the 8:2 train/validation split released by EventBind~\cite{zhou2023clip}, which also serves as the test set as the dataset provides no separate test partition.\\
For N-ImageNet-mini, we use the full $50\,\mu s$ samples, resized to $224\times224$, with the authors' train/validation splits. Validation used as test set.


\vspace{-1em}
\subsubsection{Data augmentation}
\label{ssec:data_augm}

Event data augmentation follows~\cite{li2022neuromorphicdataaugmentationtraining} with \preCR{temporal flipping, 2D spatial rolling (up to $\pm 20$ px), rotation (up to $\pm 30 \degree{}$), horizontal shear (up to $\pm 0.2$ factor) and cutout (up to 16 px side length), applied on N-Caltech101
with a probability of 0.3,
and
spatial horizontal flipping and 2D spatial rolling (up to $\pm 20$ px) applied on the larger N-ImageNet-mini with a probability of 0.5.}
%



\vspace{-1em}
\subsubsection{Full-precision baseline training}
\label{ssec:baselineCNN-train}

\preCR{Training settings are chosen empirically using a grid search strategy, to optimize accuracy and avoid overfitting.}
The ResNet-18 baseline is trained for $200$ epochs ($100$ on N-ImageNet-mini) with cross-entropy on the labels and Adam optimizer, a learning rate of $3\times10^{-3}$, a weight decay of $10^{-4}$ and batch size of 128. We use a cosine-annealed learning rate with a 5-epoch linear warmup.

\vspace{-1em}
\subsubsection{BNNs two-step training and knowledge distillation}
\label{ssec:two-steps}

Each BNN follows a two-step schedule~\cite{martinez2020trainingbinaryneuralnetworks}.
We run $200$ epochs per step on N-Caltech101 and $100$ epochs per step on N-ImageNet-mini to reduce computation time, at an initial learning rate of $10^{-2}$.
Learning rate is instead set to $10^{-3}$ for A\&B~BNN's pretrained runs to limit optimization collapse.
We did not use Adaptive Gradient Clipping from A\&B~BNN throughout: enabling it lowered accuracy on the representations and did not resolve the collapse.
Following~\cite{liu2021adamtrainingstrategieshelp}, ReActNet and A\&B~BNN use Adam with a step-1 weight decay of $5\times10^{-6}$, BNext~Small uses AdamW with a differentiated decay ($10^{-3}$ for real-valued and $10^{-8}$ for binary parameters).
The three BNNs are trained with knowledge distillation from an EventBind teacher~\cite{zhou2023clip} we trained ourselves (Section~\ref{ssec:architectures}).
We followed the distillation technique~\cite{hinton2015distillingknowledgeneuralnetwork} with temperature scaling setting of $T^2=2$, without mixing ground-truth labels. All runs use a cosine-annealed learning rate with a 5-epoch linear warmup. All models are trained with a batch size of 128, except BNext for N-ImageNet-mini, where we used 64 to fit within the same 32GB memory.

\vspace{-1em}
\subsubsection{Cross-modal Pretraining}
\label{ssec:pretraining}

We adopt a direct weight warm-start initialization from RGB pretraining as experienced in~\cite{kim2022nimagenetrobustfinegrainedobject} rather than feature alignment or distillation. 
Hence, models then skip the two-step schedule and train in a single binary stage ($100$ epochs) warm-started from RGB ImageNet weights: for 2-channel inputs we drop the first convolution's green channel, and for 6-channel inputs we duplicate the three RGB channels.
ResNet-18 is likewise initialized from ImageNet and trained for $100$ epochs.

\subsection{Classification results}
\label{ssec:classification_results}


\begin{table}[htb]
\caption{Classification accuracy (\%) on N-Caltech101 validation set and Optimal Time
Window (ms) Across all architectures and representations, from scratch and
\emph{pretrained} (ImageNet-initialised).
Bold values indicate each model's best-performing representation (scratch and pretrained considered separately).
$\Delta$ = pretrained $-$ scratch, each at its own
optimal time window.}
\label{tab:all_models_comparison}
\centering
\begin{adjustbox}{width=1\textwidth}
\begin{tabular}{l l c c c | c}
\hline
& & \multicolumn{3}{c|}{\textbf{Binary Architectures}} & \textbf{Full Precision Baseline} \\
\textbf{Representation} & & \textbf{A\&B~BNN} & \textbf{BNext} & \textbf{ReActNet} & \textbf{ResNet-18 (FP)} \\
\hline
\multirow{3}{*}{EventCount} & scratch    & 81.73 (100 ms) & 83.34 (20 ms)  & 83.34 (50 ms)  & 81.45 (100 ms) \\
                            & pretrained & 84.55 (20 ms) & 89.09 (100 ms)            & 87.48 (100 ms)            & 85.41 (20 ms) \\
                            & $\Delta$   & +2.82          & +5.74                     & +4.14                     & +3.96 \\
\hline
\multirow{3}{*}{SAE}        & scratch    & 79.78 (300 ms) & 83.11 (50 ms)  & 82.60 (200 ms) & 82.25 (200 ms) \\
                            & pretrained & 85.81 (100 ms)            & 89.20 (200 ms)            & 87.36 (300 ms)            & 85.64 (300 ms) \\
                            & $\Delta$   & +6.03                     & +6.09                     & +4.77                     & +3.39 \\
\hline
\multirow{3}{*}{PEV6}       & scratch    & \textbf{83.34 (200 ms)} & \textbf{84.84 (300 ms)} & \textbf{84.20 (300 ms)} & \textbf{83.46 (200 ms)} \\
                            & pretrained & \textbf{86.33 (200 ms)}            & \textbf{90.58 (300 ms)}            & \textbf{88.91 (300 ms)}            & \textbf{86.33 (300 ms)} \\
                            & $\Delta$   & +2.99                              & +5.74                              & +4.71                              & +2.87 \\
\hline
\multirow{3}{*}{BEI (binary)}        & scratch    & 77.25 (50 ms)  & 82.42 (50 ms)  & 81.85 (20 ms)  & 81.05 (50 ms) \\
                            & pretrained & 83.69 (20 ms) & 87.82 (20 ms)             & 85.18 (20 ms)             & 84.61 (50 ms) \\
                            & $\Delta$   & +6.44         & +5.40                     & +3.33                     & +3.56 \\
\hline
\multirow{3}{*}{PBEV6 (binary)}      & scratch    & 80.87 (20 ms) & 82.54 (50 ms)  & 81.96 (200 ms) & 82.77 (300 ms) \\
                            & pretrained & 80.30 (300 ms) & 88.45 (50 ms)             & 86.79 (50 ms)            & 85.41 (50 ms) \\
                            & $\Delta$   & $-$0.57        & +5.92                     & +4.82                    & +2.64 \\
\hline
\end{tabular}
\end{adjustbox}
\end{table}

\begin{figure}[htb]
    \centering
    \begin{minipage}[t]{0.32\linewidth}\centering
        \includegraphics[width=\linewidth]{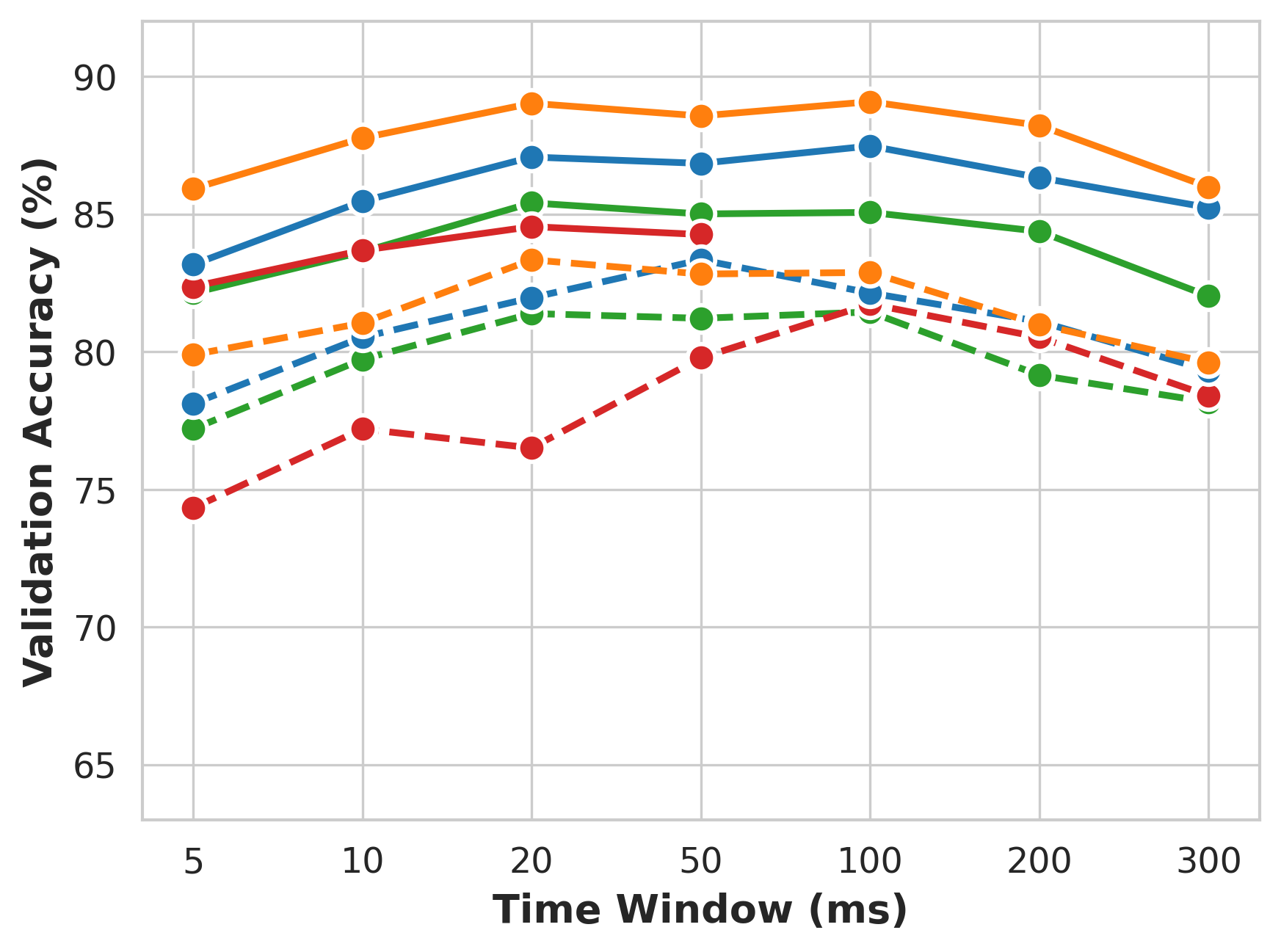}\\(a)~Event Count
    \end{minipage}\hfil
    \begin{minipage}[t]{0.32\linewidth}\centering
        \includegraphics[width=\linewidth]{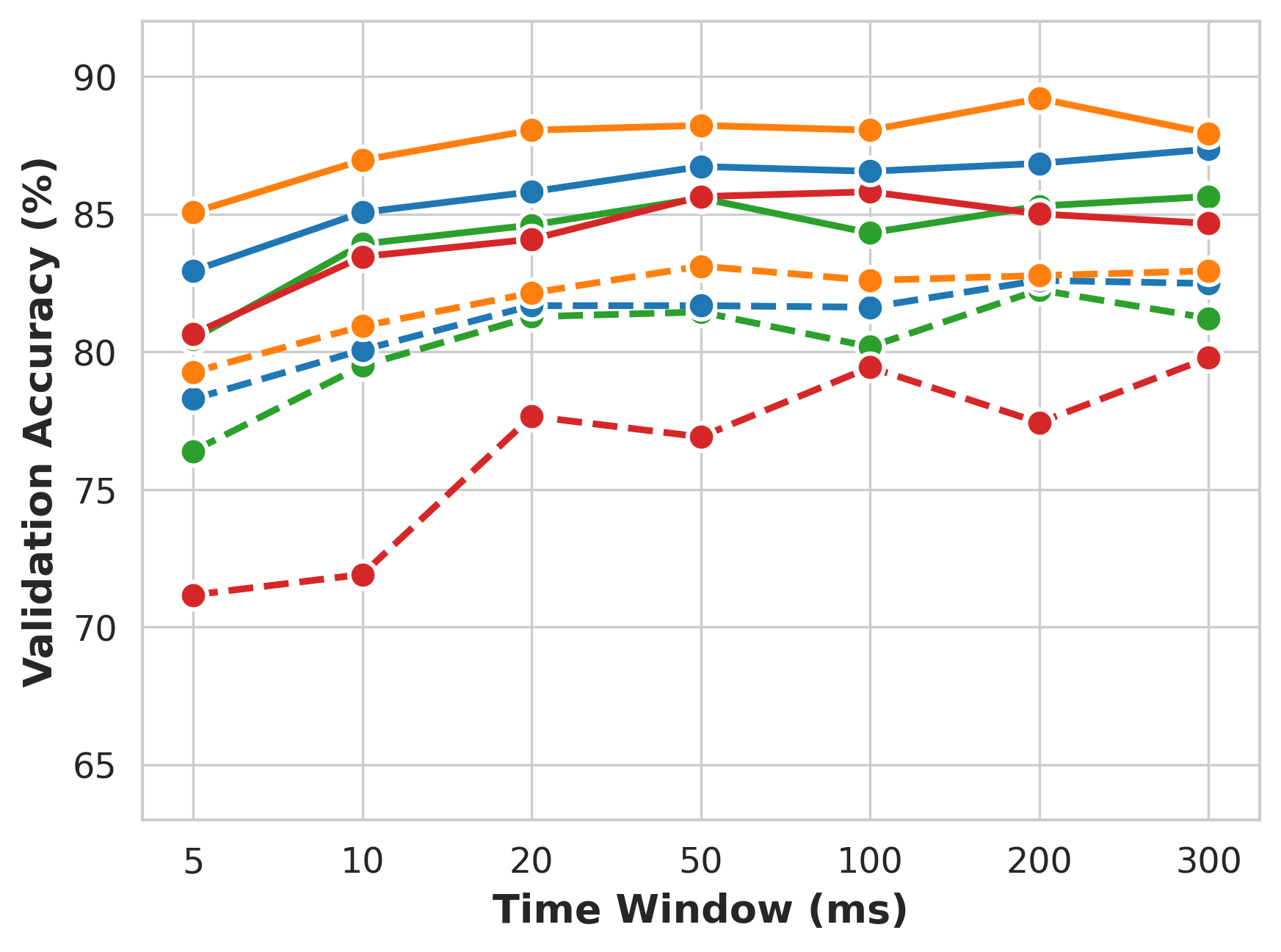}\\(b)~SAE
    \end{minipage}\hfil
    \begin{minipage}[t]{0.32\linewidth}\centering
        \includegraphics[width=\linewidth]{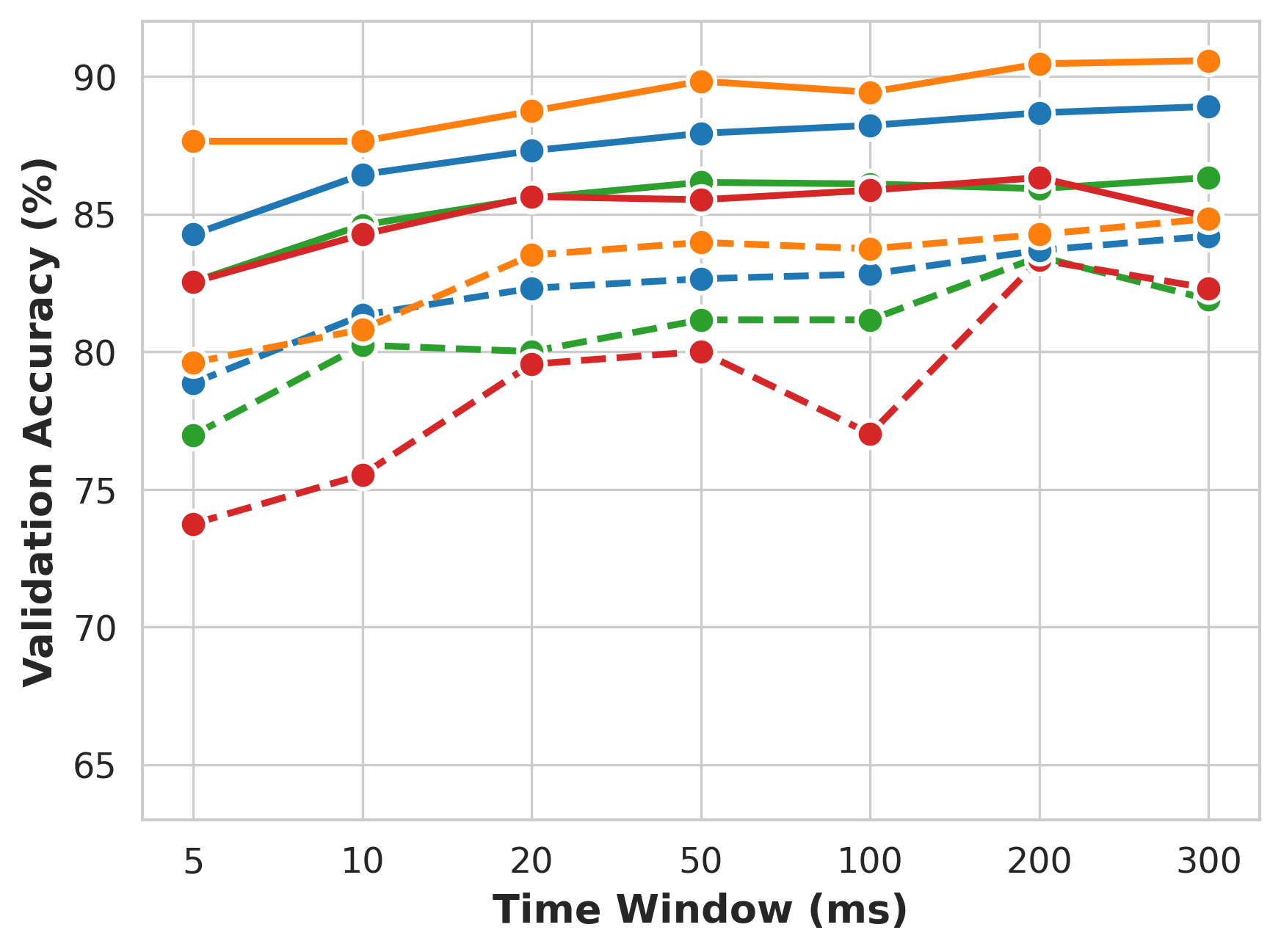}\\(c)~PEV6
    \end{minipage}

    \vspace{0.2cm} 
    
    \begin{minipage}[t]{0.32\linewidth}\centering
        \includegraphics[width=\linewidth]{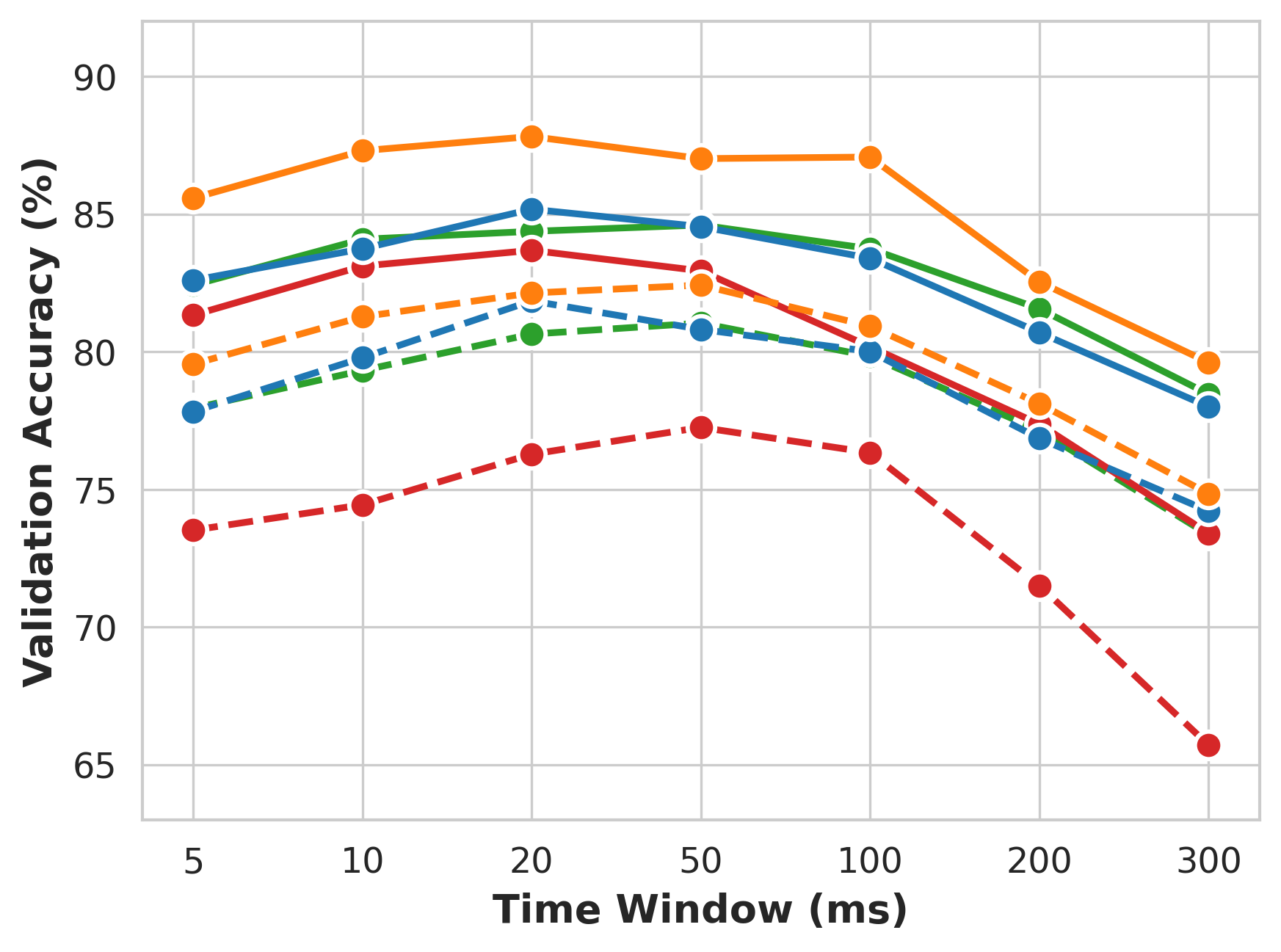}\\(d)~BEI (binary)
    \end{minipage}\hfil
    \begin{minipage}[t]{0.32\linewidth}\centering
        \includegraphics[width=\linewidth]{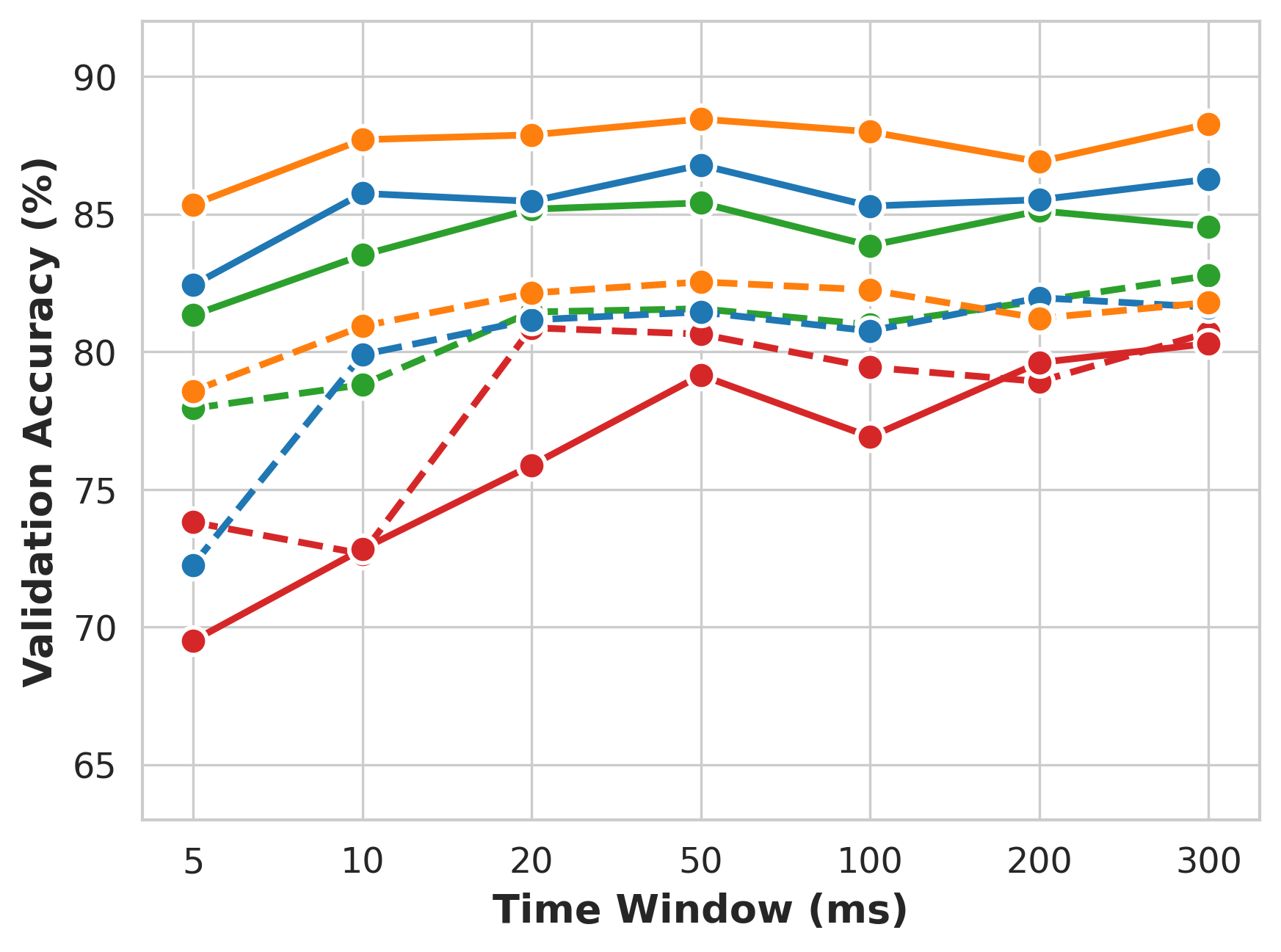}\\(e)~PBEV6 (binary)
    \end{minipage}
    
    \vspace{0.2cm}
    
    \includegraphics[width=0.95\linewidth]{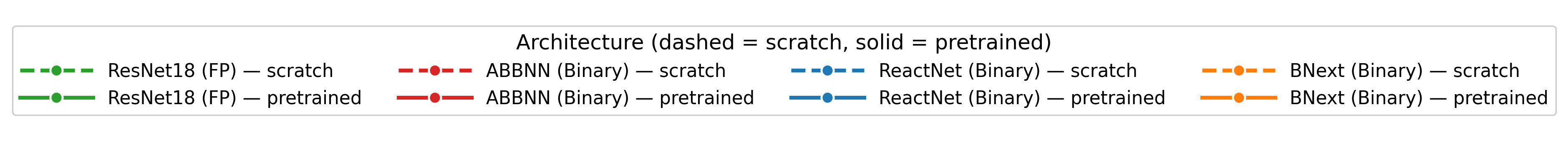}

    \caption{Validation N-Caltech101 set accuracy vs.\ time window for various event representations. The colour mapping denotes the different network architectures.}
    \label{fig:combined_acc_vs_time}
\end{figure}

\begin{table}[htb]
\caption{N-ImageNet-mini validation accuracy (\%), from scratch and
\emph{pretrained} (ImageNet-initialised) models. Bold values indicate best
representation per architecture (scratch and pretrained considered
separately). $\Delta$ = pretrained $-$ scratch. \\ \preCR{$^{\dagger}$~denotes a failed training.}
}
\label{tab:nimagenet_mini}
\centering
\begin{adjustbox}{width=0.85\textwidth}
\begin{tabular}{l l c c c | c}
\toprule
& & \multicolumn{3}{c|}{\textbf{Binary Architectures}} & \textbf{Full Precision Baseline} \\
\textbf{Representation} & & \textbf{A\&B~BNN} & \textbf{BNext} & \textbf{ReActNet} & \textbf{ResNet-18 (FP)} \\
\midrule
\multirow{3}{*}{EventCount} & scratch    & 39.62          & \textbf{47.86} & \textbf{40.02} & \textbf{43.96} \\
                             & pretrained & \preCR{1.00$^{\dagger}$} & \textbf{50.16} & \textbf{48.38} & \textbf{46.94} \\
                             & $\Delta$   & $-$38.62       & +2.30          & +8.36          & +2.98 \\
\midrule
\multirow{3}{*}{SAE}        & scratch    & \textbf{44.74} & 43.62 & 39.74 & 41.32 \\
                             & pretrained & \textbf{46.20} & 48.62 & 46.84 & 45.64 \\
                             & $\Delta$   & +1.46          & +5.00 & +7.10 & +4.32 \\
\midrule
\multirow{3}{*}{PEV6}       & scratch    & 37.06   & 36.96  & 35.36  & 28.14 \\
                             & pretrained & 41.24   & 49.78  & 48.06  & 36.82 \\
                             & $\Delta$   & +4.18   & +12.82 & +12.70 & +8.68 \\
\midrule
\multirow{3}{*}{BEI (binary)}        & scratch    & 35.04   & 41.00 & 36.20 & 39.74 \\
                             & pretrained & 41.86   & 45.50 & 44.54 & 43.16 \\
                             & $\Delta$   & +6.82   & +4.50 & +8.34 & +3.42 \\
\midrule
\multirow{3}{*}{PBEV6 (binary)}      & scratch    & 36.80   & 37.70 & 36.30 & 29.32 \\
                             & pretrained & 40.90   & 45.22 & 46.20 & 33.10 \\
                             & $\Delta$   & +4.10   & +7.52 & +9.90 & +3.78 \\
\bottomrule
\end{tabular}
\end{adjustbox}
\end{table}

Results for chosen architectures (Section~\ref{ssec:architectures}) and event representations (Section~\ref{ssec:event_repr}) are given in Table~\ref{tab:all_models_comparison} (N-Caltech101) and Table~\ref{tab:nimagenet_mini} (N-ImageNet-mini).
\preCR{To simplify, they show only the best accuracy across accumulation times for each setup, that would be elected for deployment without latency consideration.}
\preCR{As a complement}, Fig.~\ref{fig:combined_acc_vs_time} illustrates the \preCR{full} N-Caltech101 performance trajectories.

\vspace{-1em}
\subsubsection{Accuracy when trained from scratch}

As the accumulation window increases, the models ingest more contextual information, often leading to higher classification accuracy on N-Caltech101 (Table~\ref{tab:all_models_comparison}), but representations differ in effectively using that extra time, \preCR{as illustrated in Figure~\ref{fig:combined_acc_vs_time}}.
The BEI, that discards events timing and density, saturates early:
it peaks at short windows (best 82.42\% at 50\,ms with BNext, and as early as 20\,ms for ReActNet) and then degrades.
\preCR{But overall, it offers the lowest accuracies.}
Conversely, the 6 bins temporal volume PEV6, \preCR{able to keep most of events timing and density}, keeps improving up to the full extract, reaching the overall best result for every backbone
(84.84\% at 300\,ms with BNext).
EventCount, PBEV6 and SAE, \preCR{all dropping either the events density or timing}, sit between these extremes, each within roughly two points of the PEV6 optimum at their own best time window (83.34\% at 20ms, 82.54\% at 50ms and 83.11\% at 50ms respectively, all with BNext).
\preCR{Unlike PEV6, the related PBEV6 representation stops improving with the longest accumulation times.}
For a single shot, end-to-end latency is the accumulation window plus one forward pass, so this window choice sets the latency.

Comparing N-Caltech101 and N-ImageNet-mini (Table~\ref{tab:nimagenet_mini}), this ranking inverts.
\preCR{For the N-ImageNet-mini very short samples duration, the 6 bins representations not only loose their advantages in time encoding, they also drastically reduce the final accuracies.}
PEV6, while being the best encoder for every backbone on N-Caltech101, collapses on N-ImageNet-mini, most severely for ResNet-18 whose accuracy falls to 28.14\% \preCR{(the worst of all)}.
The dense two-channels EventCount and SAE are the strongest representations for N-ImageNet-mini, achieving respectively 47.86\% and 43.62\% accuracy with BNext, again the best backbone.
\preCR{This shows that for ultra short accumulation times (\textit{i.e.} sparse samples as sown in Figure~\ref{fig:sae_cross_dataset}), we shall consider to simplify the input representation and drop events timing or density, allowing for denser channels.}
Compared to  N-Caltech101, the lower best scratch accuracies illustrate how N-ImageNet-mini is harder to process.
Distilled binary networks already match or exceed the full-precision baseline from scratch on both datasets, with BNext leading on both.

\vspace{-1em}
\subsubsection{Effect of cross-modal pretraining}
Fig.~\ref{fig:combined_acc_vs_time} shows pretrained models surpass scratch training for every time window, representation and architecture except A\&B~BNN with PBEV6.
Tables~\ref{tab:all_models_comparison} and~\ref{tab:nimagenet_mini} report the gain $\Delta$ per representation and architecture, each at its own optimal window: binary networks exploit the pretrained initialization more than the full-precision baseline does.
On N-Caltech101, BNext benefits the most and most consistently ($+5.40$ to $+6.09$), ahead of ReActNet ($+3.33$ to $+4.82$), itself ahead of the full-precision ResNet-18 baseline ($+2.64$ to $+3.96$). A\&B~BNN pretrained on EventCount is the exception: it collapses at long accumulation windows, falling to 27.63\%
and below from 100ms.
Its architecture uses scaled weight standardization rather than batch normalization, so once pretrained it has no activation normalization to absorb EventCount's raw magnitude, which grows with the accumulation window. 

N-ImageNet-mini shows similar gains, with a bigger step up for PEV6 \preCR{and PBEV6,} and higher relative benefits for ReActNet than BNext. 
A\&B~BNN gains are lower and it collapses again with EventCount. 
\preCR{These rankings are consistent with the architectures' designs. Binarization lets all three BNNs afford deeper networks than ResNet-18 at a fraction of its cost (Table~\ref{tab:op-mini}). Among BNNs, BNext extends and invests over twice ReActNet's parameters and operations, while A\&B~BNN trades accuracy and training stability for hardware-oriented simplifications, notably the removal of batch normalization.}

\vspace{-1em}
\subsubsection{Binary representations: accuracy and efficiency}

Binary inputs are competitive with most state-of-the-art methods on N-Caltech101 (Table~\ref{tab:sota}), with the proposed PBEV6 slightly ahead of BEI (Table~\ref{tab:all_models_comparison}).
\preCR{PBEV6 with BNext (88.45\%) or ReActNet (86.79\%) surpasses the best full-precision ResNet-18 result (best at 86.33\% with PEV6) at an $8.6\times$ efficiency gain.}
\preCR{PBEV6 does not behave exactly as PEV6 on longer time windows, preventing it to reach as good accuracies (in Figure~\ref{fig:combined_acc_vs_time}, PBEV6 looses accuracy from 100ms). Despite this, from 10 to 50\,ms accumulation times, PBEV6 results are equivalent to SAE and EventCount and closer to PEV6, while allowing for binary input layers, making it the best choice when shorter latencies are required.}
On N-ImageNet-mini, however, the very short samples coded as PBEV6 do not carry enough information to match the best representations (SAE and EventCount).
Binary inputs remain attractive because they are extremely cheap to generate, let the binary networks themselves stay lighter and efficient with short time windows, \preCR{even lighter than the 2-channels non-binary representations such as SAE and EventCount} (Table~\ref{tab:op-mini}).
They suit constrained hardware and gives a favorable accuracy/complexity trade-off (Fig.~\ref{fig:acc_vs_ops}).

\subsection{Initialization retention analysis}
\label{ssec:init_retention}

From scratch, each binary network's final signs are uncorrelated with its own random initialization. Table~\ref{tab:c2i} shows every architecture sitting at the $0.5$ floor on both datasets.
Training therefore fully discards and is independent from the random starting point.
With RGB pretraining, final signs remain correlated with the initialization.
Table~\ref{tab:c2i} shows C2I rising to $0.73$--$0.88$ for every architecture on both datasets, well above the scratch floor. A\&B~BNN's collapsed N-Caltech101 cells depart from this range instead, reaching $0.995$: training there leaves the backbone frozen at its pretrained starting point rather than adjusting it.

\begin{table}[htb]
\centering
\caption{C2I ratio of the binary weights, (mean$\pm$std over the run grid). A\&B~BNN's N-ImageNet-mini pretrained mean excludes EventCount (failed run).}
\label{tab:c2i}
\begin{adjustbox}{width=0.6\textwidth}
\begin{tabular}{l cc cc}
\toprule
& \multicolumn{2}{c}{N-ImageNet-mini} & \multicolumn{2}{c}{N-Caltech101} \\
\cmidrule(lr){2-3}\cmidrule(lr){4-5}
Architecture & scratch & pretrained & scratch & pretrained \\
\midrule
A\&B~BNN & $0.500{\scriptstyle\,\pm0.000}$ & $0.726{\scriptstyle\,\pm0.002}$ & $0.500{\scriptstyle\,\pm0.004}$ & $0.875{\scriptstyle\,\pm0.004}$ \\
BNext & $0.504{\scriptstyle\,\pm0.002}$ & $0.768{\scriptstyle\,\pm0.002}$ & $0.508{\scriptstyle\,\pm0.002}$ & $0.830{\scriptstyle\,\pm0.002}$ \\
ReActNet & $0.500{\scriptstyle\,\pm0.000}$ & $0.780{\scriptstyle\,\pm0.000}$ & $0.500{\scriptstyle\,\pm0.000}$ & $0.774{\scriptstyle\,\pm0.002}$ \\
\bottomrule
\end{tabular}
\end{adjustbox}
\end{table}
\subsection{Theoretical computation time} 
\label{ssec:complexity-analysis}


We do not measure on-device latency in this work, but
we report the existing Xilinx ZCU 102 FPGA board implementation of BNext that sustains 72.6 FPS~\cite{ldfbnext}, that is an estimated $\approx$14 ms per inference.
According to Table~\ref{tab:op-mini}, BNext~Small is the heaviest BNN compared in our benchmark, so we can expect even faster inference for ReActNet and A\&B~BNN.

We report also the GCN EFGCN~\cite{JEZIOREK2026103850}, which processes events as point clouds and measures between 4.4 to 9.3\, ms per inference on a ZCU104 SoC FPGA.
This performance comes at a real accuracy cost, reaching between 62.8 to 64.1\% accuracy on N-Caltech101, 20 points below the BNNs at similar 50\,ms accumulation windows.
For real implementations, pseudo-frame representations competitive at short time windows are attractive for low latency neuromorphic vision systems.

\section{Conclusion}
\label{sec:conclusion}
We show that modern Binary Neural Networks can match or exceed full-precision accuracy on event-camera data while substantially cutting computational cost, when trained with knowledge distillation from a strong teacher and cross-modal pretraining. Under this regime, BNNs outperform ResNet-18 on N-Caltech101 (90.58\% accuracy with a non-binary 6-channel input at $7.5\times$ fewer operations, up to $21\times$ fewer for our binary-input configurations) and likewise outperform the full-precision baseline on the harder N-ImageNet-mini. Our Polar-wise Binary Event Volume and fully binary input blocks extend binarization to the input layer, letting a six-channel representation require fewer operations than a conventional two-channel input on an unmodified BNN. We further find that the best event representation depends primarily on the sensor's event-density and accumulation regime rather than on architecture alone, and that RGB-pretrained weights offer a useful prior that binary networks retain and exploit.

However, our evaluation is limited to classification on pan-tilt recordings of static scenes, with complexity estimated theoretically rather than measured on target hardware.
Future work should \preCR{focus on natural event recordings,} measure on-device latency and energy consumption, \preCR{further explore the capabilities of binary input representations}, improve the robustness and generalization of binary models, and extend the approach to tasks such as object detection. Together, these are steps toward low-latency, resource-efficient neuromorphic vision.


\vspace{-1em}
\section*{Acknowledgment}

\preCR{This research is part of the REVE-BNN project, funded by the French National Agency (ANR-24-CE33-4001), and was provided with computing HPC and storage resources by GENCI at IDRIS thanks to the grant 2025-AD011014065R2 on the supercomputer Jean Zay's V100 partition.}



\bibliographystyle{splncs04} 
\bibliography{biblio}

\end{document}